\pdfoutput=1

\documentclass{article}
\usepackage[utf8]{inputenc}
\usepackage[T1]{fontenc}\usepackage[sfdefault,scaled=.85]{FiraSans}\usepackage{newtxsf}
\usepackage[protrusion=true, expansion]{microtype}
\usepackage[margin=15mm, bottom=20mm]{geometry}
\usepackage{multicol}
\usepackage{graphicx}
\usepackage{amssymb}
\usepackage{amsmath}
\usepackage{multirow}
\usepackage[hidelinks,
  pdfauthor={Tomasz Tarkowski},
  pdftitle={Genetic algorithm formulation and tuning with use of test functions},
  pdfkeywords={algorithm tuning, genetic algorithm, test function}
]{hyperref}
\usepackage{bookmark}

\begin{document}

\begin{center}

  {\Large\bf Genetic algorithm formulation and tuning with use of test functions}

  \bigskip
  
  {\large \href{https://orcid.org/0000-0002-6071-9389}{Tomasz Tarkowski}}

  \medskip

  {\small\it Chair of Complex Systems Modelling, Institute of Theoretical Physics, \\
    Faculty of Physics, University of Warsaw, Pasteura 5, PL-02093 Warszawa, Poland}

\end{center}

\begin{abstract}
  This work discusses single-objective constrained genetic algorithm with floating-point, integer, binary and permutation representation.
  Floating-point genetic algorithm tuning with use of test functions is done and leads to a parameterization with comparatively outstanding performance. \hskip\stretch{1} {\scriptsize Copyright (c) 2022 Tomasz Tarkowski. License: CC BY-NC-ND 4.0 (\url{http://creativecommons.org/licenses/by-nc-nd/4.0/}).}
\end{abstract}

\section*{Introduction}
\addcontentsline{toc}{section}{Introduction}

The idea of application of biological evolution \cite{Darwin1859} and genetic \cite{Mendel1866} principles to the optimization problems was first sketched by Alan Turing in his 1948 essay titled \emph{Intelligent Machinery} \cite{Turing1948} and later extended by himself \cite{Turing1950} to a technique which is now called \emph{genetic programming} \cite{Koza1990}. These works were the very beginning of the \emph{evolutionary computations} (EC)---field of computer science devoted to the population-based, trial-and-error methods of problem solving. One of the first EC performed on a computer were done by Nils A. Barricelli in 1953 at the Institute for Advanced Study at Princeton, NJ \cite{Barricelli1962} on a machine built by John von Neumann's group \cite{Galloway2011}. Nowadays, EC field consists of many subfields---one of them is \emph{genetic algorithm} (GA) approach \cite{Holland1975}, subject of this work.

\section{Genetic algorithm}

\subsection{Genotype and its representation}

\emph{Genotype} $g$ is a finite polymorphic list of Boolean, real or integer values called \emph{genes} \cite{Fraser1957}. More precisely, $g \in X_0 \times \dots \times X_{c-1} = \prod_{i=0}^{c-1} X_i$, where $X_i$ is equal to $\mathbb{B} = \{{\rm false}, {\rm true}\}$ or is bounded subset of set of real numbers $\mathbb{R}$ or integer numbers $\mathbb{Z}$. This work considers subsets of kind of intervals of type $[a, b]$ or $[a, b]_{\mathbb{Z}} \equiv \left\{k \in \mathbb{Z} \mid a \leq k \leq b \right\}$. Despite the fact that \emph{a priori} every gene in genotype can be of different type, it is common practice to employ genotypes with pure \emph{representation}: \emph{binary} ($g\in\mathbb{B}^c$), \emph{floating-point} ($g\in\mathbb{R}^c$) or \emph{integer} ($g\in\mathbb{Z}^c$) \cite{Zhang2017}. Otherwise, the representation is called \emph{mixed} and is out of scope of this work.

Proper genotype for given optimization problem can have some additional constraints, $g \in G \subseteq \prod_{i=0}^{c-1} X_i$. The $G$ set can be arbitrary subset of $\prod_{i=0}^{c-1} X_i$, i.e. constraints imposed on different genes can be different and constraints for given gene can depend on values of all other genes. However, if $G = X_0^c$, then genotype is called \emph{uniform}. Moreover, $G$ can be defined as an extension of predicate $Q\colon \prod_{i=0}^{c-1} X_i \rightarrow \mathbb{B}$, i.e. $G = \{(x_0, \dots , x_{c-1}) \in \prod_{i=0}^{c-1} X_i \mid Q(x_0, \dots , x_{c-1}) \}$. One can define \emph{permutation representation}, where $G = \{(x_0, \dots , x_{c-1}) \in \{ 0, \dots, c - 1\}^c \mid \forall i \neq j\colon x_i \neq x_j \}$, i.e. it is extension of permutation condition predicate.

Value $c$, i.e. domain dimension of $\prod_{i=0}^{c-1} X_i$, is called \emph{genotype length} (or \emph{size}), while position in genotype is called \emph{locus} (pl. \emph{loci}), so if genotype is of length $c$ then \emph{locus} belongs to the set $\{ 0, \dots , c - 1\} \equiv \iota_c$ (notation inspired by the APL language \cite{Iverson1962}).

\subsection{Population and sequence of populations}

Sequence of genotypes of length $\mu$:
\begin{eqnarray}
\left( g_i \right)_{i=0}^{\mu - 1} \in \left( \iota_\mu \rightarrow \prod_{i=0}^{c-1} X_i \right) \equiv P_{X_0 , \dots , X_{c-1}}^\mu \subset \bigcup_{\mu =0}^{+\infty} P_{X_0, \dots , X_{c-1}}^\mu \equiv P_{X_0 , \dots , X_{c-1}}^*
\end{eqnarray}
is called \emph{population} or, in context of evolutionary step, \emph{generation} while $P_{X_0 , \dots , X_{c-1}}^\mu$ and $P_{X_0 , \dots , X_{c-1}}^*$ are sets of all possible populations of genotypes formulated basing on domain $\prod_{i=0}^{c-1} X_i$ with length $\mu$ or arbitrary (including zero), respectively. Zero-length population is marked with the symbol $\epsilon$.

Notion of population is insufficient to describe the evolutionary process, though---it is necessary to make one step further and define \emph{sequence of populations}:
\begin{eqnarray}
\left(\left( g_{i, j} \right)_{i=0}^{\mu (j) - 1}\right)_{j=0}^{\nu - 1} \in \left( \iota_\nu \rightarrow P_{X_0, \dots , X_{c-1}}^* \right) \equiv P_{X_0, \dots , X_{c-1}}^{*\nu} \subset \bigcup_{\nu = 0}^\infty P_{X_0, \dots , X_{c-1}}^{*\nu} \equiv P_{X_0, \dots , X_{c-1}}^{**} \\
\left(\left( g_{i, j} \right)_{i=0}^{\mu - 1}\right)_{j=0}^{\nu - 1} \in \left( \iota_\nu \rightarrow P_{X_0, \dots , X_{c-1}}^\mu \right) \equiv P_{X_0, \dots , X_{c-1}}^{\mu\nu} \subset \bigcup_{\nu = 0}^\infty P_{X_0, \dots , X_{c-1}}^{\mu\nu} \equiv P_{X_0, \dots , X_{c-1}}^{\mu*} \\
P_{X_0, \dots , X_{c-1}}^{\mu\nu} \subset P_{X_0, \dots , X_{c-1}}^{*\nu} \\
P_{X_0, \dots , X_{c-1}}^{\mu*} \subset P_{X_0, \dots , X_{c-1}}^{**}
\end{eqnarray}
where $P_{X_0, \dots , X_{c-1}}^{\mu\nu}$ is set of all possible sequences of length $\nu$ of populations of length $\mu$. Replacement of $\mu$ or $\nu$ to the star symbol ($*$) means all possible values of given parameter, e.g. $P_{X_0, \dots , X_{c-1}}^{*\nu}$ is set of all sequences of length $\nu$ of populations of arbitrary length.

Notation $P_{X_0, \dots , X_{c-1}}^*$ and $P_{X_0, \dots , X_{c-1}}^{**}$ are inspired by Kleene closure, but it is not equivalent to it, because--- according to Kleene algebra \cite{Kleene1951}---double application of Kleene star is equivalent to single application and here $P_{X_0, \dots , X_{c-1}}^{*} \neq P_{X_0, \dots , X_{c-1}}^{**}$, i.e. population concatenation does not occur automatically. To concatenate populations one can use flatten (\emph{flat}, i.e. Italian \emph{bemolle}) $\flat \colon P_{X_0, \dots , X_{c-1}}^{**} \rightarrow P_{X_0, \dots , X_{c-1}}^*$ function:
\begin{equation}
\flat \left( \left( g_{i,j} \right)_{i=0}^{\mu (j) - 1} \right)_{j=0}^{\nu - 1}
=
\left(
g_{0, 0}, \dots , g_{\mu (0) - 1, 0}
, \dots , \dots , \dots ,
g_{0, \nu - 1}, \dots , g_{\mu (\nu - 1) - 1, \nu - 1}
\right)
\end{equation}

\subsection{Fitness function}

Set of genotypes is by itself not very useful if there is no objective function to optimize over it, or---to put it differently---if there is no optimization problem to solve. In GA field, the term \emph{fitness function} is used and common practice is to define the problem in such a way, that this function is maximized. Fitness function, that is $f\colon G \rightarrow \mathbb{R}$, represents given maximization problem, but it does not need necessarily be formulated explicitly and calculation of its values might be costly. Fitness function ensures gradual enhancement of population ``quality'' in evolutionary process. In \emph{multi-objective GA} the fitness function is extended to the so-called \emph{cost function} $G \rightarrow \mathbb{R}^d$, where $d \in \mathbb{N}_+$ and this function is able to maximize many, often competing, parameters simultaneously and solution of such problem has form of set called \emph{Pareto frontier} \cite{Fonseca1993}. This work, however, is devoted to $d = 1$ case.

\subsection{Variation operators}

The essence of metaheuristics is intelligent search of space of potential solutions with intention of finding the best one. In order to achieve that in GA approach the mechanism of exploitation of successful genotypes' genes is applied. This mechanism includes \emph{variation operator} dependent on representation, which is a function $v_{n,m}\colon V_{X_0, \dots , X_{c-1}}^{n, m}$, where $V_{X_0, \dots , X_{c-1}}^{n, m} \equiv P_{X_0, \dots , X_{c-1}}^n \rightarrow P_{X_0, \dots , X_{c-1}}^m$. Variation $v_{n,m}$ applied on \emph{parents} $\left( g_i \right)_{i=0}^{n-1}$ results in \emph{offspring} $\left( h_i \right)_{i=0}^{m-1}$, where every element is called \emph{child}. Application of variation on population will be denoted with $v_{n,m} \left( g_i \right)_{i=0}^{n-1}$ or similar.

There are three main cases of variation: \emph{mutation} (or \emph{unary variation}) $v_{1,1}$ and two \emph{recombinations} (or \emph{binary variations})---$v_{2,1}$ (one child) and $v_{2,2}$ (two children). Variations of other kinds are less often used and are out of the scope of this work. Variations can be composed and are often used that way. For case of recombination and mutation the canonical composition is to first apply recombination and then the mutation. Mutation is in that case, using term taken from functional programming,  \emph{mapped} on every child obtained from the recombination process, i.e. $\left( {\rm map}_n v_{1,1}\right) \circ v_{2,n}$, where in special case of operations on populations ${\rm map}_n \colon V_{X_0, \dots , X_{c-1}}^{1, 1} \rightarrow V_{X_0, \dots , X_{c-1}}^{n, n}$, i.e. ${\rm map}_\mu v_{1, 1} \left( g_i \right)_{i=0}^{\mu - 1} = \left( v_{1, 1} \left( g_i \right) \right)_{i=0}^{\mu - 1}$.

\subsection{Probability distributions}

Variations are often based on drawing procedure from some probability distribution. In this work normal $\mathcal{N}$, uniform $\mathcal{U}$ and special case of Bernoulli distributions ${\rm B}$ will be used---their definitions can be easily found in literature. Normal distribution with mean $\mu$ and standard deviation $\sigma$ is marked here $\mathcal{N}(\mu , \sigma )$ (contrary to conventional form $\mathcal{N}(\mu , \sigma^2 )$). Uniform distribution has continuous form $\mathcal{U}(a, b)$ for $a, b \in \mathbb{R}$, where values are drawn from closed interval $[a, b]$, and discrete form  $\mathcal{U}\left\{ a, b \right\}$ for $a, b \in \mathbb{Z}$ or $a, b \in \mathbb{B}$, where values are drawn from set $[a, b]_{\mathbb{Z}}$ or $\left\{a, b\right\} \subset \mathbb{B}$, respectively. Furthermore, the following notation for drawing from set $X$ is also assumed:
\begin{equation}
\mathcal{U} (X) = \left\{
\begin{array}{lcl}
  \mathcal{U}(a, b) & \Leftrightarrow & X = [a, b] \\
  \mathcal{U}\{ a, b\} & \Leftrightarrow & X = [a, b]_{\mathbb{Z}} \vee X = \left\{ a, b \right\} \subset \mathbb{B} \\
\end{array} \right.
\end{equation}
For Bernoulli distribution ${\rm B}(1, p)$ the probability of drawing value $1$ (equivalent to ${\rm true} \in \mathbb{B}$) equals to $p$. Value $0$ (equivalent to ${\rm false} \in \mathbb{B}$) can be drawn with probability $1 - p$.

\subsection{Examples of mutation operators}

\begin{itemize}
\item \emph{Gaussian mutation} is a floating-point variation having two parameters: $\sigma \in \mathbb{R}_+$ and $p \in [0, 1]$. Variation of each gene in a given genotype occurs with probability $p$ and consists of addition of a value $\sigma \cdot \mathcal{N}(0, 1)$. In case where such mutation was to move \emph{locus} $i$ gene value out of constraints imposed by $X_i$ set, the $\inf X_i$ or $\sup X_i$ value is used instead.
\item \emph{Swap mutation} is an uniform genotype variation, where for genotype of size $c$ it consists of swapping values of two genes with \emph{loci} drawn from $\mathcal{U}(\iota_c)$.
\item \emph{Random-reset mutation} is a binary, floating-point and integer variation having one parameter $p \in [0, 1]$. Each gene of a given genotype is changed with probability $p \in [0, 1]$ and variation of gene \emph{locus} $i$ consists of assigning new value drawn from $\mathcal{U}(X_i)$.
\end{itemize}

\subsection{Examples of recombination operators}

\begin{itemize}
\item \emph{Arithmetic recombination} is a floating-point variation of type $V_{X_0, \dots , X_{c-1}}^{2,1}$ and when applied to $( x_i )_{i=0}^{c-1}$ and $( x_i' )_{i=0}^{c-1}$ then result consists of one child equal to $( (x_i + x_i') / 2 )_{i=0}^{c-1}$.
\item \emph{Single arithmetic recombination} is a floating-point variation of type $V_{X_0, \dots , X_{c-1}}^{2,2}$ and when applied to $( x_i )_{i=0}^{c-1}$ and $( x_i' )_{i=0}^{c-1}$ then result consists of two children equal to $( x_0 , \dots , ( x_k + x_k' ) / 2, \dots , x_{c-1} )$ and $( x_0' , \dots , \frac{1}{2} ( x_k + x_k' ) , \dots , x_{c-1}' )$, where \emph{locus} $k$ is drawn from $\mathcal{U}(\iota_c)$.
\item \emph{One-point crossover} is a binary, floating-point and integer variation of type $V_{X_0, \dots , X_{c-1}}^{2,2}$, where \emph{locus} $k$ is drawn from $\mathcal{U}(\iota_c)$ and genotypes $(x_0, \dots , x_{k-1} , x_k' , \dots , x_{c-1}' )$ and $(x_0', \dots , x_{k-1}' , x_k , \dots , x_{c-1} )$ are obtained, i.e. ``tails'' of parent genotypes $( x_i )_{i=0}^{c-1}$ and $( x_i' )_{i=0}^{c-1}$ are exchanged. This variation can be easily extended to $n$-point crossover.
\item \emph{Cut-and-crossfill recombination} is a permutation variation of type $V_{X_0, \dots , X_{c-1}}^{2,2}$, where \emph{locus} $k$ is drawn from $\mathcal{U}\left( \iota_c \setminus \left\{ 0 \right\}\right)$, first $k$ genes are copied from first parent to first child and, analogously, from second parent to second child, then genotype of first child is filled with not yet used genes of second parent in order of increasing \emph{loci} and, analogously, second child is filled with genes from first parent.
\end{itemize}

\subsection{Self-adaptive mutation}

Self-adaptive mutation \cite{Back2000} is an extension of Gaussian mutation, where standard deviation $\sigma$ also evolves and is \emph{de facto} part of the genotype. There are several types of self-adaptive mutation, but only the most popular version will be shown. It employs $c$ additional genes containing values of $\sigma_i$ on each direction of optimization problem, i.e. instead of $g = (x_0, \dots , x_{c-1}) \in \prod_{i=0}^{c-1} X_i$ and $G$, genotype $g_{\sigma} = (x_0, \dots , x_{c-1}, \sigma_0, \dots , \sigma_{c-1}) \in \prod_{i=0}^{c-1} X_i \times \prod_{i=0}^{c-1} \Sigma_i$ and $G_{\Sigma}$, where $\Sigma_i \subset \mathbb{R}_+$, are used. Genotype size is here equal to $c_{\sigma} = 2c$, domain $G_{\Sigma}$ can be defined as predicate extension as well, while the self-adaptive mutation is operator of class $V_{X_0, \dots , X_{c-1}, \Sigma_0 , \dots , \Sigma_{c-1}}^{1,1}$. Extension of the fitness function from $f\colon G \rightarrow \mathbb{R}$ to $f\colon G_{\Sigma} \rightarrow \mathbb{R}$ is trivial, because $\sigma_i$ does not influence $f$ values.

Contrary to Gaussian mutation each gene is mutated unconditionally (with probability equal to $1$) for self-adaptive mutation and the process itself has two stages. First, every $\sigma_i$ gene is mutated: $\sigma_i' = \sigma_i \cdot \exp \left( \tau_0 \cdot \mathcal{N}(0, 1) + \tau_1 \cdot \mathcal{N}_i (0, 1) \right)$, where $\mathcal{N}(0, 1)$ is drawn once while $\mathcal{N}_i (0, 1)$ is drawn for every gene $\sigma_i$. Next, every $x_i$ is mutated with use of new values of $\sigma_i'$ according to the formula $x_i' = x_i + \sigma_i' \cdot \mathcal{N}_i (0, 1)$. If mutation of gene $\sigma_i$ or $x_i$ was to move gene value out of constraints then, likewise in Gaussian mutation, infimum or supremum value is used instead. Self-adaptive mutation has two parameters, $\tau_0$ and $\tau_1$, which values depend on genotype length:
\begin{equation}
\tau_0 \propto 1 / \sqrt{2c}, \quad \tau_1 \propto 1 / \sqrt{2\sqrt{c}}
\end{equation}

\subsection{Predicate $Q$ violations}

Result of variation $v_{n,m} \left( g_i \right)_{i=0}^{n-1} = \left( h_i \right)_{i=0}^{m-1}$ for some combination of $v_{n,m}$, $\left( g_i \right)_{i=0}^{n-1}$ and $Q$ chosen by GA practitioner might violate predicate $Q$, i.e. $\exists i \in \iota_m \colon \neg Q h_i$, which is equivalent to $h_i \notin G$ for some $i$. From point of view of logic it might be considered as an imperfection of problem formulation. On the other hand, at some occasions, the cost of creation of new variation operator proper for the given predicate might outweigh time overhead resulting from slower algorithm. Proposed solution is to modify the fitness function so it would be able to treat problematic $h_i$. Previously defined fitness function $f\colon G \rightarrow \mathbb{R}$ can be extended to the whole domain, $f_Q \colon \prod_{i=0}^{c-1} X_i \rightarrow \mathbb{R}$:
\begin{equation}
  f_Q (g) = \left\{
  \begin{array}{ll}
    f(g) & g \in G \\
    -\Delta_f & g \in \prod_{i=0}^{c-1} X_i \setminus G
  \end{array}
  \right.
\end{equation}
where $-\Delta_f$ is chosen such, that $\forall g \in G \colon -\Delta_f \ll f(g)$, i.e. improper genotype has small chance of selection to the next generation and to the multiset of parents. This extension can be done provided that $\min_{g \in G} f(g) > -\infty$, which is reasonable assumption in numerical calculations. Unfortunately, $\min_{g \in G} f(g)$ might not be known \emph{a priori}, so for the sake of simplicity the fitness function can be modified even further taking $-\Delta_f = -\infty$, i.e. $f_Q \colon \prod_{i=0}^{c-1} X_i \rightarrow \mathbb{R} \cup \{ -\infty \}$.

\subsection{Stochastic variation operator}

Beside that variation operators are often stochastic \emph{per se}, for purpose of GA, \emph{stochastic variation operator} $S\colon V_{X_0, \dots , X_{c-1}}^{n, m} \times [0, 1] \rightarrow V_{X_0, \dots , X_{c-1}}^{n, m}$ is defined in such a way, that it randomly decides whether variation provided as its argument is applied or not:
\begin{equation}
S \left( v_{n,m}, p \right) \left( g_i \right)_{i=0}^{n-1}
=
\left\{
\begin{array}{ll}
  v_{n,m}\left( g_i \right)_{i=0}^{n-1} & {\rm if}\ s \\
  i_{n,m}\left( g_i \right)_{i=0}^{n-1} & {\rm if}\ \neg s \\
\end{array}
\right.
{\rm for}\ s = {\rm B}(1, p)
\end{equation}
where $i_{n,m}\colon V_{X_0, \dots , X_{c-1}}^{n, m}$ is defined for $n, m \geq 0$:
\begin{eqnarray}
  i_{n,n} \left( g_i \right)_{i=0}^{n-1} = \left( g_0, \dots , g_{n-1} \right) \phantom{,} & \\
  i_{n,n+1} \left( g_i \right)_{i=0}^{n-1} = \left( g_0, \dots , g_{k-1} , g_k , g_k, g_{k+1} , \dots , g_{n-1} \right) , &
  k = \mathcal{U}(\iota_n ) \\
  i_{n,n-1} \left( g_i \right)_{i=0}^{n-1} = \left( g_0, \dots , g_{k-1} , g_{k+1} , \dots , g_{n-1} \right) , &
  k = \mathcal{U}(\iota_n )
\end{eqnarray}
i.e. sequence is expanded (shortened) through copy (deletion) of randomly selected elements according to uniform probability distribution.

\subsection{Selection}

The second part of intelligent search of space of all possible solutions is genotype \emph{selection}. Selection is a function $p_{\mu\nu}^{\pi}\colon P_{X_0, \dots , X_{c-1}}^{*\nu} \rightarrow P_{X_0, \dots , X_{c-1}}^\mu$ and there are two important cases: \emph{parent selection} ($\nu = 1$) and \emph{survivor selection} ($\nu = 2$), which---for the purpose for this work---is called \emph{selection to the next generation}. Every algorithm of class $\nu = 1$ can be easily generalized into arbitrary $\nu > 0$ with composition $p_{\mu\nu}^{\pi} = p_{\mu 1}^{\pi} \circ \flat$. This composition will be useful, especially in case of stochastic universal sampling mechanism described further.

Selection algorithm $p_{\mu\nu}^{\pi}$ is parameterized with \emph{selection probability function} \cite{Holland1975} $\pi\colon P_{X_0, \dots , X_{c-1}}^{\mu '} \rightarrow [0, 1]^{\mu '}$, which determines selection probabilities of each genotype in a given population and which satisfies condition:
\begin{equation}
\sum_{i=0}^{\mu ' - 1} \left. \pi \left( g_0, \dots , g_{\mu ' - 1} \right) \right|_i = 1
\end{equation}

As an example of selection probability function \emph{fitness proportional selection} (a.k.a. fitness \emph{proportionate} selection) with \emph{windowing} procedure (FPS) was chosen:
\begin{equation}
  \left. \pi_{\rm FPS} \left( g_i \right)_{i=0}^{\mu ' - 1} \right|_{i_0} = \left\{
  \begin{array}{ll}
    \displaystyle \frac{f(g_{i_0}) - \min_{j \in \iota_{\mu '} \wedge g_j \in G} f(g_j) + 1 / \mu '}{1 - \mu ' \cdot \min_{j \in \iota_{\mu '} \wedge g_j \in G} f(g_j) + \sum_{j \in \iota_{\mu '} \wedge g_j \in G} f(g_j)} & g_{i_0} \in G \\
    0 & g_{i_0} \in \prod_{i=0}^{c-1} X_i \setminus G
  \end{array}
  \right.
\end{equation}
This function is well defined if $\exists i \in \iota_{\mu '} \colon g_i \in G$, otherwise optimization problem must be reformulated as there are too many violations of $Q$. In an extreme case of population size $\mu '$ of equally fit genotypes, e.g. $(g_0, \dots , g_0)$, this function returns $\left(1 / \mu ', \dots , 1 / \mu ' \right)$, which is the expected result.

The aforementioned FPS mechanism has insufficient selection pressure in some applications, though. If for two genotypes $g_0$, $g_1$ one of them is slightly more fit, $f(g_0) = f(g_1) + \varepsilon_f$, then also $\left. \pi_{\rm FPS} \left( g_i \right)_i \right|_0 \approx \left. \pi_{\rm FPS} \left( g_i \right)_i \right|_1$, i.e. there is practically no preference of $g_0$ over $g_1$ during selection. Therefore, \emph{ranking selection} \cite{Baker1987} will be introduced.

Let us, however, define several helper functions first. The $F_P\colon V_{X_0, \dots , X_{c-1}}^{*,*}$ function filters the population according to a given predicate $P\colon \prod_{i=0}^{c-1} X_i \rightarrow \mathbb{B}$:
\begin{equation}
F_P \left( g_i \right)_{i=0}^{\mu' - 1} = \left( g_{i_k} \right)_{i_k \in \left\{ j \in \iota_{\mu'} \mid P(g_j) \right\} },\quad i_0 < \dots < i_{\mu'' - 1},\quad \mu'' = \# \left\{ j \in \iota_{\mu'} \mid P(g_j) \right\}
\end{equation}
where $\#$ stands here for cardinality of set and will shortly be reused with new meaning. The $\#\colon P_{X_0, \dots , X_{c-1}}^* \rightarrow \mathbb{N}$ function returns size of the population, i.e. $\# \left( g_i \right)_{i=0}^{\mu' - 1} = \mu'$. Obviously, $\# \circ F_P \left( g_i \right)_{i=0}^{\mu' - 1} = \# \{ j \in \iota_{\mu'} \mid P(g_j) \}$. The last helper function, $\sigma_{f_Q}\colon V_{X_0, \dots , X_{c-1}}^{\mu' , \mu'}$, performs stable sort of the population according to the ascending fitness function values $f_Q$:
\begin{equation}
\sigma_{f_Q} \left( g_i \right)_{i=0}^{\mu' - 1} = \left( g_{i_k} \right)_{i_k \in \iota_{\mu'}},\quad f_Q (g_{i_0}) \leq \dots \leq f_Q (g_{i_{\mu' - 1}}),\quad f_Q (g_{i_j}) = f_Q (g_{i_{j+1}}) \Rightarrow i_j < i_{j+1}
\end{equation}

Finally, ranking selection (RS) can be defined as:
\begin{equation}
\pi_{\rm RS} \left( h_i \right)_{i=0}^{\mu ' - 1} = \flat \left( \left( 0 \right)_{j = 0}^{\mu_{\neg Q}' - 1} , \left( r_{\mu_Q' , j} \right)_{j = 0}^{\mu_Q' - 1} \right) , \quad \left( h_i \right)_{i=0}^{\mu ' - 1} = \sigma_{f_Q} \left( g_i \right)_{i=0}^{\mu ' - 1}
\end{equation}
where $\mu_{\neg Q}' = \# F_{\neg Q} \left( g_i \right)_{i=0}^{\mu' - 1}$, $\mu_Q' = \# F_{Q} \left( g_i \right)_{i=0}^{\mu' - 1}$ and $\mu' = \mu_{\neg Q}' + \mu_Q'$, while $r_{\mu_Q' , j}$ is selection probability with linear or exponential pressure:
\[
r_{\mu_Q' , j}^{\rm lin} = \left\{
\begin{array}{l}
  1 \\
  \frac{2 - s}{\mu_Q'} + \frac{2j (s - 1)}{\mu_Q' (\mu_Q' - 1)}, \, 1 < s \leq 2
\end{array}
\right.
, \quad
r_{\mu_Q' , j}^{\rm exp} = \left\{
\begin{array}{ll}
  1 & {\rm if}\ \mu_Q' = 1 \\
  \frac{1 - e}{\mu_Q' (1 - e) + e - e^{1 - \mu_Q'}} \cdot \left( 1 - e^{-j} \right) & {\rm if}\ \mu_Q' > 1
\end{array}
\right.
\]
Similarly to the FPS, ranking selection is well defined if at least one genotype in the population being its argument satisfies predicate $Q$. Contrary to the FPS, equally fit genotypes are given different selection probabilities with RS.

\subsection{Examples of selection $p_{\mu\nu}^{\pi}$}
\begin{itemize}
\item \emph{Roulette wheel algorithm} (RWA) is of class $p_{1, 1}^{\pi}$. This mechanism is traditionally explained with use of wheel with one arm and with fields of angular width proportional to selection probability function $\pi$ value for given genotype. Drawing genotype from this algorithm can be compared to spinning the wheel and drawn genotype is pointed out by the arm. In order to draw $\mu$ genotypes one has to perform $\mu$ algorithm runs.
\item \emph{Stochastic universal sampling} (SUS) \cite{Baker1987} is of class $p_{\mu 1}^{\pi}$ and is extension of RWA with $\mu$ equidistant arms where draw of $\mu$ genotypes occurs in one run.
\item \emph{Generational selection} (GS) is trivial algorithm of class $p_{\mu 2}$ where from two populations of equal sizes (current generation and offspring) returns second (offspring).
\end{itemize}

\subsection{Genetic algorithm}

GA starts with some initial population $\left( h_i \right)_{i=0}^{\mu - 1} \in P_{X_0, \dots , X_{c-1}}^\mu$. This population can be selected \emph{ad hoc} or randomly with some probability distribution---the only requirement is that the use of procedure $v_{0, \mu}\colon V_{X_0, \dots , X_{c-1}}^{0, \mu}$, which creates first generation $v_{0, \mu} \epsilon = \left( h_i \right)_{i = 0}^{\mu - 1}$, should guarantee that $\forall i\colon h_i \in G$. If random procedure was used, e.g. $h = \left( \mathcal{U}(X_i) \right)_{i=0}^{c-1}$, then afterwards $h$ should be rejected if $\neg Q(h)$ and procedure should be repeated. Selection of bounded sets in definition of gene stated previously is not accidental, because otherwise drawing from uniform distribution would violate Kolmogorov probability axioms \cite{Kolmogorov1950}.

After initial population selection, the algorithm performs loop, where next generation is created based on previous one. Next generation creation process starts with parent selection $p_{k \cdot n, 1}^{\pi}$. Parents population is divided into tuples of size $n$ and for every tuple variation $v_{n,m}$ is applied, resulting with offspring size $m$. Total offspring can be marked as $\left( h_i \right)_{i=0}^{\lambda - 1} \in P_{X_0, \dots , X_{c-1}}^\lambda$, where $\lambda = k \cdot m$. Then, selection to the next generation $p_{\mu 2}^{\pi}$ is applied, i.e. selection of $\mu$ genotypes from previous generation of size $\mu$ and offspring of size $\lambda$. Sequence of populations generated during evolutionary process consisting of initial population and populations generated through selection to the next generation is called \emph{evolution}. In GA it is common approach to use constant size of generation over whole evolutionary process, so if generation size equals to $\mu$ then evolution of size $\nu$ is an element of $P_{X_0, \dots , X_{c-1}}^{\mu\nu}$ while evolution of unknown size belongs to $P_{X_0, \dots , X_{c-1}}^{\mu *}$. It is easy to note, that evolution of type $P_{X_0, \dots , X_{c-1}}^{\mu *}$ is Markov chain with discrete time \cite{Aarts1989}.

Evolutionary loop stops when \emph{termination condition} in form of predicate $\mathbb{Z} \times P_{X_0, \dots , X_{c-1}}^{\mu *} \rightarrow \mathbb{B}$, taking loop counter and evolution produced so far, is fulfilled. Termination conditions can be joined with conjunction or disjunction.

\subsection{Examples of termination condition}

\begin{itemize}
\item Reaching of maximum number of permitted iterations.
\item Reaching of \emph{plateau} of fitness function.
\item Reaching previously specified value of fitness function by any genotype.
\end{itemize}

\subsection{GA extensions}

An \emph{abstract} GA was introduced here alongside with concrete example realizations of its constituents. However, there are some extensions to the basic algorithm, e.g. introducing spatial structure, i.e. \emph{cellular} GA \cite{Alba2008} (being special case of cellular automaton \cite{vonNeumann1966, Cerruti2020}), where genotypes are vertices of some connected graph and can recombine only with their neighbors. These extensions are out of scope of this work.

\subsection{GA implementation}

For the purpose of this work custom C++ implementation named Quilë available at \url{https://github.com/ttarkowski/quile/} was used.

\section{Optimization algorithm benchmark---test functions}

\subsection{Test functions. Algorithm tuning}

Optimization algorithms for problems of $\mathbb{R}^c \rightarrow \mathbb{R}^d$ type, including floating-point GA, can be benchmarked with use of so-called \emph{test functions} (TFs). The aim is to evaluate the performance of optimum finding capabilities of given algorithm or its parameterization---Pareto frontier in multi-objective optimization or point in $\mathbb{R}^c$ space in ordinary single-objective algorithm. Different parameterizations of one algorithm (e.g. genetic) can be compared with each other with use of TFs---this procedure can be used for algorithm \emph{tuning} in order to increase its performance. Here, floating-point single-objective GA effectivity and efficiency analysis with use of TFs will be presented.

It is common practice that TF $f^* \colon \prod_{i=0}^{c-1} X_i \rightarrow \mathbb{R}$ is minimized, i.e. one searches for point $\vec{x}_{\min} \in \prod_{i=0}^{c-1} X_i \subset \mathbb{R}^c$ such, that $\forall \vec{x}\in \prod_{i=0}^{c-1} X_i \colon f^* (\vec{x}_{\min}) \leq f^* (\vec{x})$. Point $\vec{x}_{\min}$ is also denoted as $\arg \min_{\vec{x}} f^* (\vec{x})$. From the numerical point of view function minimization relies on finding such approximation of minimum $\vec{x}_{{\rm approx} \min}$, which satisfies two conditions. Firstly, obviously, this approximation should be close to real minimum (``proximity in domain''), i.e. $| \vec{x}_{{\rm approx} \min} - \vec{x}_{\min} | \leq \varepsilon_{\vec{x}}$ for some small $\varepsilon_{\vec{x}}$. Secondly, function value at approximation point should approximate function value at real minimum (``proximity in codomain''), i.e. $| f^* ( \vec{x}_{{\rm approx} \min} ) - f^* ( \vec{x}_{\min} ) | \leq \varepsilon_{f^*}$ for some small $\varepsilon_{f^*}$. Both conditions are not equivalent---one can consider multimodal function with nearly deep minima to show that second condition does not imply the first one and unimodal function with discontinuity points around real minimum to show that first condition does not imply the second one.

History of research on optimization problems, not only floating-point, delivers substantial set of TFs (problems). These are scattered around different scientific reports and compiled into repositories and review articles of different size and quality (\emph{caveat emptor}). One of the positively standing out resource in regard to size, quality, documentation and ease of use is still developed MINLPLib repository (\url{http://www.minlplib.org/}), which contains problems of binary, integer, floating-point and mixed types with different complexity of objective function and predicate defining its domain formulated algebraically (complexity of type linear, quadratic, polynomial and signomial \cite{Peterson1966}).

\begin{figure}
\centering
\includegraphics{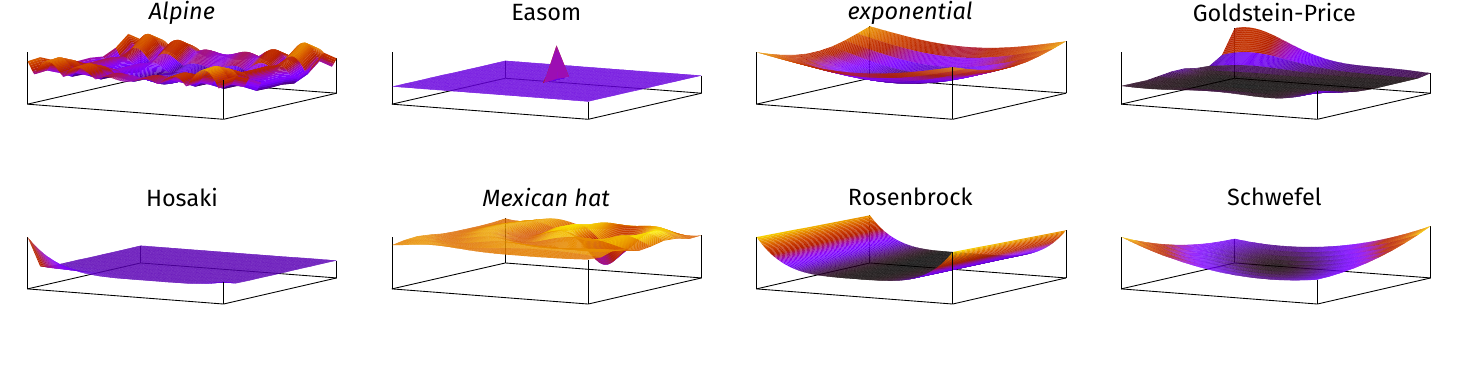}
\caption{Selected test functions for 2-dimensional domain.}
\label{fig1}
\end{figure}

For the purpose of this work 16 single-objective TFs were selected (Ackley, \emph{Alpine}, Aluffi-Pentini, Booth, Colville, Easom, \emph{exponential}, Goldstein-Price, Hosaki, Leon, Matyas, \emph{Mexican hat}, Miele-Cantrell, Rosenbrock, Schwefel, \emph{sphere}) from literature \cite{Ackley1987, AluffiPentini1985a, AluffiPentini1985b, Easom1990, Goldstein1971, Rosenbrock1960, Schwefel1981, Jamil2013, Jamil2013a}. These TFs are defined in Tab.~\ref{tab1}, visualized for selected cases in Fig.~\ref{fig1}, implemented in Quilë library and are not necessarily contained in aforementioned MINLPLib repository.

TFs can be classified with respect to continuity, convexity, codomain dimensionality (single- and multi-objective), domain dimensionality ($c$ value), number of local minima (uni- and multimodal), separability or using descriptive terms (e.g. ``valleys'', ``basins''). Separability occurs when:
\begin{equation}
\arg \min_{x_0, \dots , x_{c-1}} f^* (x_0, \dots , x_{c-1}) = \left( \arg \min_{x_0} f^* (x_0, \dots ) , \dots , \arg \min_{x_{c-1}} f^* (\dots , x_{c-1}) \right)
\end{equation}

Optimization algorithm tuning process is by itself optimization task. This raises natural question, whether GA parameterization can be found using some algorithm, even genetic. The answer is positive and such genetic mechanism is called \emph{metagenetic algorithm} \cite{Grefenstette1986} while from group of other procedures one can mention e.g. \emph{F-Race algorithm} \cite{Birattari2010}. Unfortunately, none of these techniques is commonly used by EC practitioners and it will not be employed here either. The tuning process will be performed using method of testing intuitively or conventionally chosen parameters. The key point of the whole process is statistical analysis.

\begin{table}[!hb]
    \caption{Test functions implemented in Quilë library: 1.~Ackley, 2.~\emph{Alpine}, 3.~Aluffi-Pentini, 4.~Booth, 5.~Colville, 6.~Easom, 7.~\emph{exponential}, 8.~Goldstein-Price, 9.~Hosaki, 10.~Leon, 11.~Matyas, 12.~\emph{Mexican hat}, 13.~Miele-Cantrell, 14.~Rosenbrock, 15.~Schwefel, 16.~\emph{sphere}. The $x_{0, {\rm AP}}$ value was calculated to 15 decimal places: $x_{0, {\rm AP}} = \left. \frac{2 \sqrt{3}}{3} \cos \left( \frac{1}{3} \arccos \left( -\frac{3 \sqrt{3}}{2} q \right) - \frac{2 \pi}{3} k \right) \right|_{q = \frac{1}{10},\, k = 2} \approx -1{.}046680531804602$.}
    \label{tab1}
    \centering
    \begin{tabular}{|r|l|l|l|l|}

    \hline
    \# &
    $c$ &
    $f^* \left( \vec{x} \right)$ &
    $\prod_{i=0}^{c-1} X_i$ &
    $\vec{x}_{\min}$ \\

    \hline\hline

    1. &
    $n$ &
    $\displaystyle -20 \exp\left( \frac{-0{.}02}{\sqrt{n}} \sqrt{\sum_{i = 0}^{n - 1} x_i^2} \right) - \exp\left( \frac{1}{n} \sum_{i = 0}^{n - 1} \cos \left( 2 \pi x_i \right) \right) + 20 + e$ &
    $[-35, 35]^n$ &
    $(0 , \dots , 0)$ \\

    2. &
    $n$ &
    $\displaystyle \sum_{i = 0}^{n - 1} \left| x_i \sin x_i  + 0{.}1 x_i \right|$ &
    $[-10, 10]^n$ &
    $(0 , \dots , 0)$ \\

    3. &
    $2$ &
    $\displaystyle \frac{1}{4} x_0^4 - \frac{1}{2} x_0^2 + \frac{1}{10} x_0 + \frac{1}{2} x_1^2$ &
    $[-10, 10]^2$ &
    $(x_{0, {\rm AP}}, 0)$ \\

    4. &
    $2$ &
    $\displaystyle (x_0 + 2x_1 - 7)^2 + (2x_0 + x_1 - 5)^2$ &
    $[-10, 10]^2$ &
    $(1, 3)$ \\

    5. &
    $4$ &
    $\displaystyle 100 \left( x_0 - x_1^2 \right)^2 + \left( 1 - x_0 \right)^2 + 90 \left( x_3 - x_2^2 \right)^2 + \left( 1 - x_2 \right)^2$ &
    $[-10, 10]^4$ &
    $(1, 1, 1, 1)$ \\
    &
    &
    $\displaystyle +\ 10{.}1 \left( x_1 - 1 \right)^2 + \left( x_3 - 1 \right)^2 + 19{.}8 \left( x_1 - 1 \right) \left( x_3 - 1 \right)$ &
    &
    \\

    6. &
    $2$ &
    $\displaystyle -\cos x_0 \cdot \cos x_1 \cdot \exp \left( -(x_0 - \pi )^2 - (x_1 - \pi )^2 \right)$ &
    $[-100, 100]^2$ &
    $( \pi , \pi )$ \\

    7. &
    $n$ &
    $\displaystyle -\exp\left( -\frac{1}{2} \sum_{i = 0}^{n - 1} x_i^2 \right)$ &
    $[-1, 1]^n$ &
    $(0 , \dots , 0)$ \\

    8. &
    $2$ &
    $\displaystyle \left( 1 + \left( x_0 + x_1 + 1 \right)^2 \left( 19 - 14 x_0 + 3 x_0^2 - 14 x_1 + 6 x_0 x_1 + 3 x_1^2 \right) \right)$ &
    $[-2, 2]^2$ &
    $(0, -1)$ \\
    &
    &
    $\displaystyle \cdot \left( 30 + \left( 2 x_0 - 3 x_1 \right)^2 \left( 18 - 32 x_0 + 12 x_0^2 + 48 x_1 - 36 x_0 x_1 + 27 x_1^2 \right) \right)$ &
    &
    \\

    9. &
    $2$ &
    $\displaystyle \left( 1 - 8 x_0 + 7 x_0^2 - \frac{7}{3} x_0^3 + \frac{1}{4} x_0^4 \right) x_1^2 \exp (-x_1)$ &
    $[-10, 10]^2$ &
    $(4, 2)$ \\

    10. &
    $2$ &
    $\displaystyle 100 \left( x_1 - x_0^2 \right)^2 + \left( 1 - x_0 \right)^2$ &
    $[-1{.}2, 1{.}2]^2$ &
    $(1, 1)$ \\

    11. &
    $2$ &
    $\displaystyle 0{.}26 \left( x_0^2 + x_1^2 \right) - 0{.}48 x_0 x_1$ &
    $[-10, 10]^2$ &
    $(0, 0)$ \\

    12. &
    $2$ &
    $\displaystyle -20 \frac{\sin g(x_0, x_1)}{g(x_0, x_1)}, \, g(x_0, x_1) = 0{.}1 + \sqrt{(x_0 - 4)^2 + (x_1 - 4)^2}$ &
    $[-10, 10]^2$ &
    $(4, 4)$ \\

    13. &
    $4$ &
    $\displaystyle \left( \exp (-x_0) - x_1 \right)^4 + 100 \left( x_1 - x_2 \right)^6 + \tan^4 (x_2 - x_3) + x_0^8$ &
    $[-1, 1]^4$ &
    $(0, 1, 1, 1)$ \\

    14. &
    $n$ &
    $\displaystyle \sum_{i = 0}^{n - 2} \left( 100 \left( x_{i + 1} - x_i^2 \right)^2 + \left( x_i - 1 \right)^2 \right)$ &
    $[-30, 30]^n$ &
    $(1, \dots , 1)$ \\

    15. &
    $n$ &
    $\displaystyle \sum_{i = 0}^{n - 1} \left( \sum_{j = 0}^i x_i \right)^2$ &
    $[-100, 100]^n$ &
    $(0, \dots , 0)$ \\

    16. &
    $n$ &
    $\displaystyle \sum_{i = 0}^{n - 1} x_i^2$ &
    $[0, 10]^n$ &
    $(0 , \dots , 0)$ \\

    \hline

    \end{tabular}
\end{table}

\subsection{Statistical parameters}

Statistical analysis of optimization algorithms performance is done for fixed parameterization---for each TF the series of minimization attempts is performed in order to obtain statistical sample. For purpose of performance description one can use several statistical parameters connected to the number of successfully finished optimization attempts, average number of fitness function or $Q$ predicate evaluations, average total number of generated unique genotypes or average ``best'' genotype's fitness function value at given moments of the algorithm \cite{Craenen}. The following parameters were here used:
\begin{itemize}
\item For description of algorithm's \emph{effectivity} standard SR (success rate) parameter was used. It is defined as fraction of successfully finished (i.e. optimum was found) search processes to the total number of processes.
\item \emph{Efficiency} is described by AUS and $\sigma_{\rm AUS}$ parameters equal to \underline{a}verage number of \underline{u}nique individuals to get a \underline{s}olution (i.e. in successfully finished search processes) and standard deviation (root of the unbiased estimator of variance) corresponding to the aforementioned average, respectively. The AUS parameter was designed specifically for purpose of this work.
\item Description of quality of minimum approximation found by the algorithm was done with average distance between function value at real minimum and its approximation $\overline{|\Delta f^*|}$ and with average distance between real minimum and its approximation $\overline{|\Delta \vec{x}|}$. Furthermore, corresponding standard deviations $\sigma_{\overline{|\Delta f^*|}}$ and $\sigma_{\overline{|\Delta \vec{x}|}}$ were also employed.
\end{itemize}

Given the fact, that Quilë library uses database of calculated fitness function values and that these values are computed once for each unique genotype, parameters AUS and $\sigma_{\rm AUS}$ should be good metric of stochastic algorithm complexity in case of objective function, which is costly to calculate, i.e. its computation time is of the order of magnitude of seconds or more.

\begin{table}
\caption{Maximum optimization effectivity (SR) realized by some GA parameterization ($2k$, $p_{\rm r}$, $p_{\rm m}$ and---in Gaussian mutation case---$r$) for $\varepsilon_{f^*} = 10^{-1}$, $\varepsilon_{\vec{x}} = 10^{-2}$ precision. Please see \emph{examples/benchmark/results\_detailed.txt} file in Quilë repository for more details.}
\label{tab2}
\centering
\medskip
\begin{tabular}{|c|c|r|r|r|r|r|r|r|r|r|r|r|r|r|r|r|r|r|}
\hline
 & $\pi$ & c & \rotatebox{90}{Ackley\phantom{o}} & \rotatebox{90}{\emph{Alpine}\phantom{o}} & \rotatebox{90}{Aluffi-Pentini\phantom{o}} & \rotatebox{90}{Booth\phantom{o}} & \rotatebox{90}{Colville\phantom{o}} & \rotatebox{90}{Easom\phantom{o}} & \rotatebox{90}{\emph{exponential}\phantom{o}} & \rotatebox{90}{Goldstein-Price\phantom{o}} & \rotatebox{90}{Hosaki\phantom{o}} & \rotatebox{90}{Leon\phantom{o}} & \rotatebox{90}{Matyas\phantom{o}} & \rotatebox{90}{\emph{Mexican hat}\phantom{o}} & \rotatebox{90}{Miele-Cantrell\phantom{o}} & \rotatebox{90}{Rosenbrock\phantom{o}} & \rotatebox{90}{Schwefel\phantom{o}} & \rotatebox{90}{\emph{sphere}\phantom{o}} \\ \hline
\hline
\multirow{9}{*}{\rotatebox{90}{Gaussian m., arithm. r.}} & \multirow{3}{*}{\rotatebox{90}{FPS}} & 2 & $100$ & $56$ & $99$ & $90$ & -- & $15$ & $100$ & $99$ & $0$ & $45$ & $100$ & $100$ & -- & $46$ & $100$ & $100$ \\
 & & 4 & $7$ & $4$ & -- & -- & $0$ & -- & $100$ & -- & -- & -- & -- & -- & $27$ & $0$ & $72$ & $100$ \\
 & & 8 & $0$ & $0$ & -- & -- & -- & -- & $0$ & -- & -- & -- & -- & -- & -- & $0$ & $0$ & $100$ \\
 & \multirow{3}{*}{\rotatebox{90}{lin-RS}} & 2 & $100$ & $56$ & $100$ & $100$ & -- & $100$ & $100$ & $100$ & $0$ & $100$ & $100$ & $100$ & -- & $82$ & $98$ & $100$ \\
 & & 4 & $0$ & $2$ & -- & -- & $0$ & -- & $100$ & -- & -- & -- & -- & -- & $100$ & $0$ & $0$ & $100$ \\
 & & 8 & $0$ & $0$ & -- & -- & -- & -- & $0$ & -- & -- & -- & -- & -- & -- & $0$ & $0$ & $100$ \\
 & \multirow{3}{*}{\rotatebox{90}{exp-RS}} & 2 & $100$ & $100$ & $99$ & $91$ & -- & $97$ & $100$ & $100$ & $0$ & $88$ & $100$ & $100$ & -- & $67$ & $100$ & $100$ \\
 & & 4 & $4$ & $5$ & -- & -- & $0$ & -- & $100$ & -- & -- & -- & -- & -- & $35$ & $0$ & $10$ & $100$ \\
 & & 8 & $0$ & $0$ & -- & -- & -- & -- & $0$ & -- & -- & -- & -- & -- & -- & $0$ & $0$ & $100$ \\
 \hline
\hline
\multirow{12}{*}{\rotatebox{90}{Gaussian m., single arithm. r.}} & \multirow{4}{*}{\rotatebox{90}{FPS}} & 2 & $100$ & $52$ & $100$ & $91$ & -- & $32$ & $100$ & $100$ & $1$ & $31$ & $100$ & $100$ & -- & $26$ & $100$ & $100$ \\
 & & 4 & $22$ & $4$ & -- & -- & $0$ & -- & $100$ & -- & -- & -- & -- & -- & $46$ & $0$ & $4$ & $100$ \\
 & & 8 & $0$ & $0$ & -- & -- & -- & -- & $0$ & -- & -- & -- & -- & -- & -- & $0$ & $0$ & $100$ \\
 & & 16 & $0$ & $0$ & -- & -- & -- & -- & $0$ & -- & -- & -- & -- & -- & -- & $0$ & $0$ & $100$ \\
 & \multirow{4}{*}{\rotatebox{90}{lin-RS}} & 2 & $100$ & $53$ & $100$ & $99$ & -- & $100$ & $100$ & $100$ & $0$ & $90$ & $99$ & $100$ & -- & $79$ & $99$ & $100$ \\
 & & 4 & $100$ & $14$ & -- & -- & $0$ & -- & $100$ & -- & -- & -- & -- & -- & $100$ & $0$ & $1$ & $100$ \\
 & & 8 & $40$ & $0$ & -- & -- & -- & -- & $100$ & -- & -- & -- & -- & -- & -- & $0$ & $0$ & $100$ \\
 & & 16 & $0$ & $0$ & -- & -- & -- & -- & $98$ & -- & -- & -- & -- & -- & -- & $0$ & $0$ & $100$ \\
 & \multirow{4}{*}{\rotatebox{90}{exp-RS}} & 2 & $100$ & $100$ & $100$ & $87$ & -- & $84$ & $100$ & $100$ & $1$ & $33$ & $100$ & $100$ & -- & $29$ & $100$ & $100$ \\
 & & 4 & $5$ & $3$ & -- & -- & $0$ & -- & $100$ & -- & -- & -- & -- & -- & $29$ & $0$ & $3$ & $100$ \\
 & & 8 & $0$ & $0$ & -- & -- & -- & -- & $2$ & -- & -- & -- & -- & -- & -- & $0$ & $0$ & $100$ \\
 & & 16 & $0$ & $0$ & -- & -- & -- & -- & $0$ & -- & -- & -- & -- & -- & -- & $0$ & $0$ & $100$ \\
 \hline
\hline
\multirow{15}{*}{\rotatebox{90}{random-reset m., single arithm. r.}} & \multirow{5}{*}{\rotatebox{90}{FPS}} & 2 & $100$ & $47$ & $100$ & $88$ & -- & $100$ & $100$ & $100$ & $0$ & $33$ & $78$ & $100$ & -- & $33$ & $87$ & $100$ \\
 & & 4 & $100$ & $13$ & -- & -- & $0$ & -- & $100$ & -- & -- & -- & -- & -- & $1$ & $1$ & $0$ & $86$ \\
 & & 8 & $45$ & $2$ & -- & -- & -- & -- & $100$ & -- & -- & -- & -- & -- & -- & $0$ & $0$ & $4$ \\
 & & 16 & $0$ & $0$ & -- & -- & -- & -- & $57$ & -- & -- & -- & -- & -- & -- & $0$ & $0$ & $0$ \\
 & & 32 & $0$ & $0$ & -- & -- & -- & -- & $0$ & -- & -- & -- & -- & -- & -- & $0$ & $0$ & $0$ \\
 & \multirow{5}{*}{\rotatebox{90}{lin-RS}} & 2 & $100$ & $74$ & $100$ & $98$ & -- & $100$ & $100$ & $100$ & $0$ & $47$ & $54$ & $100$ & -- & $10$ & $87$ & $100$ \\
 & & 4 & $100$ & $33$ & -- & -- & $0$ & -- & $100$ & -- & -- & -- & -- & -- & $0$ & $1$ & $1$ & $100$ \\
 & & 8 & $100$ & $6$ & -- & -- & -- & -- & $100$ & -- & -- & -- & -- & -- & -- & $1$ & $0$ & $100$ \\
 & & 16 & $100$ & $0$ & -- & -- & -- & -- & $100$ & -- & -- & -- & -- & -- & -- & $0$ & $0$ & $100$ \\
 & & 32 & $100$ & $0$ & -- & -- & -- & -- & $100$ & -- & -- & -- & -- & -- & -- & $0$ & $0$ & $100$ \\
 & \multirow{5}{*}{\rotatebox{90}{exp-RS}} & 2 & $100$ & $49$ & $100$ & $90$ & -- & $99$ & $100$ & $99$ & $0$ & $38$ & $83$ & $100$ & -- & $29$ & $78$ & $100$ \\
 & & 4 & $96$ & $13$ & -- & -- & $0$ & -- & $100$ & -- & -- & -- & -- & -- & $0$ & $0$ & $1$ & $100$ \\
 & & 8 & $17$ & $1$ & -- & -- & -- & -- & $100$ & -- & -- & -- & -- & -- & -- & $0$ & $0$ & $33$ \\
 & & 16 & $0$ & $0$ & -- & -- & -- & -- & $100$ & -- & -- & -- & -- & -- & -- & $0$ & $0$ & $0$ \\
 & & 32 & $0$ & $0$ & -- & -- & -- & -- & $80$ & -- & -- & -- & -- & -- & -- & $0$ & $0$ & $0$ \\
 \hline
\end{tabular}
\end{table}

\subsection{Benchmark method. Results}

The Quilë library GA performance benchmark was done for TFs from Tab.~\ref{tab1} for $c \in \{2^{i} \mid i \in \iota_{\eta} \setminus \{ 0 \} \}$, where $\eta$ value was chosen individually for different variation operators. The recombination and mutation operators were applied stochastically with recombination probability $p_{\rm r}$ and mutation probability $p_{\rm m}$ equal to $1$ or $0{.}5$. Exploitation was done mostly through the recombination, while exploration---through mutation. Calculations were divided into three groups differentiated by variation operator:
\begin{itemize}
\item arithmetic recombination with Gaussian mutation with $p = 1 / c$, while $\sigma$ was adapted to each TF individually with formula $\sigma = r \cdot \min_{i \in \iota_c} ( b_i - a_i )$, where $[a_i, b_i] \equiv X_i$ and $r \in \{ 50\% , 5\% , 0{.}5\% \}$,
\item single arithmetic recombination with Gaussian mutation with parameters identical with the point above,
\item single arithmetic recombination with random-reset mutation with $p = 1 / c$.
\end{itemize}

Generation size $\mu$ was equal to $100$, which is relatively small number and implies low probing of space of possible solutions during creation of first random generation. Simulations were done for parent multiset of size $2k \in \{ 2, 4, 8, 16, 32, 64 \}$. Each genetic process used SUS mechanism in order to enhance quality of parent selection and selection to the next generation---FPS and RS with linear (lin-RS, $s = 2$) and exponential (exp-RS) pressure procedures were used. Absolute precision of minimum finding in codomain $\varepsilon_{f^*}$ was set to $10^{-1}$, while in domain $\varepsilon_{\vec{x}}$ to $10^{-2}$. GA was terminated when some genotype approached the real minimum to the distance of at most $\varepsilon_{f^*}$ in codomain and to the distance of at most $\varepsilon_{\vec{x}}$ in domain \emph{or} after reaching limit of $10^5$ iterations in order to stop ineffective processes. The numerical simulations were done for every possible parameter combination and for each parameterization they were performed $100$ times in order to collect appropriate statistics. The result for each parameterization and for each TF consists of SR, AUS, $\sigma_{\rm AUS}$, $\overline{|\Delta f^*|}$, $\sigma_{\overline{|\Delta f^*|}}$, $\overline{|\Delta \vec{x}|}$ and $\sigma_{\overline{|\Delta \vec{x}|}}$. The best SR values are shown in Tab.~\ref{tab2}. The detailed results can be assessed by analyzing the \emph{examples/benchmark/} directory of the Quilë library repository. For the sake of brevity only the most important conclusions will be shown further.

\subsection{Conclusions}

By analyzing the numerical simulations results one can observe several properties of described GA. Firstly, choosing each parameter value of GA process in isolation might lead to poor performance, even when every parameter might be individually proper. This parameterization applied to concrete TF might be unable to find function minimum, while other might have for this concrete TF the SR equal to $100$.

Secondly, optimization performance is obviously decreasing with problem dimensionality. It happens, because potential solution space volume grows exponentially with dimension. With the increase of $c$ value the AUS parameter increases and SR decreases. Fortunately, the AUS parameter grows slower than size of space of possible solutions, which can be assessed analyzing the best calculation series for Ackley, \emph{exponential} and \emph{sphere} TFs in case of random-reset mutation with single arithmetic recombination with lin-RS, where SR parameter was equal to $100$ for $c \in \{2, 4, 8, 16, 32\}$. Moreover, for Ackley function the best efficiency was achieved by the same parameterization ($p_{\rm m} = 0{.}5$, $p_{\rm r} = 1$, $2k = 64$) and relation ${\rm AUS} \sim c^{1{.}85 \pm 0{.}08}$ has occurred.

Thirdly, some TFs (Colville, Hosaki) were not optimized at all with chosen strategy and some were optimized with moderate efficiency. This observation is emanation of NFL theorem \cite{Wolpert1997}: GA is comparatively versatile tool, but, on the other hand, its efficiency is not high. This rule is confirmed even with apparent exception of \emph{sphere} TF for optimization with Gaussian mutation. This function has its minimum on the edge of domain, which causes that Gaussian mutation, being unable to cross the boundary, can very quickly select the point on edge, which drastically helps finding the minimum.

\section*{Summary}
\addcontentsline{toc}{section}{Summary}

Theory of genetic algorithm was discussed. Algorithm tuning with use of test functions was done and the best parameterization was found. Evolutionary computations has found wide applications in many disciplines which is proven by review literature \cite{Katoch2021, Ghaheri2015, Goudos2016, Kudjo2017, Lee2018, Drachal2021}. Genetic algorithm is the tool worth knowing.

\section*{Acknowledgments}
\addcontentsline{toc}{section}{Acknowledgments}

This work is a result of the projects funded by the National Science Centre of Poland (Twardowskiego 16, PL-30312 Kraków, Poland, \url{http://www.ncn.gov.pl/}) under the grant number \mbox{UMO-2016}\allowbreak{}\mbox{/23}\allowbreak{}\mbox{/B}\allowbreak{}\mbox{/ST3}\allowbreak{}\mbox{/03575}. 

\begin{multicols}{2}
  \bibliographystyle{elsarticle-num}
  \addcontentsline{toc}{section}{References}
  \bibliography{refs}

\begin{thebibliography}{10}
\expandafter\ifx\csname url\endcsname\relax
  \def\url#1{\texttt{#1}}\fi
\expandafter\ifx\csname urlprefix\endcsname\relax\def\urlprefix{URL }\fi
\expandafter\ifx\csname href\endcsname\relax
  \def\href#1#2{#2} \def\path#1{#1}\fi

\bibitem{Darwin1859}
C.~Darwin, {On the Origin of Species}, {John Murray}, {London, England, Great
  Britain}, 1859.

\bibitem{Mendel1866}
R.~J. Mendel, {Versuche über Pflanzenhybriden}, {Wilhelm Engelmann}, {Leipzig,
  Germany}, 1866.

\bibitem{Turing1948}
A.~M. Turing, {Intelligent Machinery}, Tech. rep., {National Physical
  Laboratory}, {Teddington, England, Great Britain} (1948).

\bibitem{Turing1950}
A.~M. Turing, {Computing Machinery and Intelligence}, {Mind} 59~(236) (1950)
  433--460.

\bibitem{Koza1990}
J.~R. Koza, {Genetic Programming: A Paradigm for Genetically Breeding
  Populations of Computer Programs to Solve Problems}, Tech. rep., {Stanford
  University Computer Science Department}, {Stanford, CA, US} (1990).

\bibitem{Barricelli1962}
N.~A. Barricelli, {Numerical testing of evolution theories}, {Acta
  Biotheoretica} 16 (1962) 69--98.
\newblock \href {https://doi.org/10.1007/BF01556771}
  {\path{doi:10.1007/BF01556771}}.

\bibitem{Galloway2011}
A.~R. Galloway, {Creative Evolution}, {Cabinet. A quarterly of art and culture}
  42 (2011) 45--50.

\bibitem{Holland1975}
J.~H. Holland, {Adaptation in Natural and Artificial Systems}, {University of
  Michigan Press}, {Ann Arbor, MI, US}, 1975.

\bibitem{Fraser1957}
A.~S. Fraser, {Simulation of Genetic Systems by Automatic Digital Computers I.
  Introduction}, {Australian Journal of Biological Sciences} 10 (1957)
  484--491.
\newblock \href {https://doi.org/10.1071/BI9570484}
  {\path{doi:10.1071/BI9570484}}.

\bibitem{Zhang2017}
L.~Zhang, H.~Pan, Y.~Su, X.~Zhang, Y.~Niu, {A Mixed Representation-Based
  Multiobjective Evolutionary Algorithm for Overlapping Community Detection},
  {IEEE Transactions on Cybernetics} 47~(9) (2017) 2703--2716.
\newblock \href {https://doi.org/10.1109/TCYB.2017.2711038}
  {\path{doi:10.1109/TCYB.2017.2711038}}.

\bibitem{Iverson1962}
K.~E. Iverson, {A Programming Language}, {John Wiley \&{} Sons, Inc.}, {New
  York, NY, US}, 1962.

\bibitem{Kleene1951}
S.~C. Kleene, {Representation of Events in Nerve Nets and Finite Automata},
  Tech. Rep. {RM-704}, {U.S. Air Force \&{} The RAND Corporation}, {Santa
  Monica, CA, US}, {Project RAND, Research Memorandum} (1951).

\bibitem{Fonseca1993}
C.~M. Fonseca, P.~J. Fleming, {Genetic Algorithms for Multiobjective
  Optimization: Formulation, Discussion and Generalization}, in: {Proceedings
  of the 5th International Conference on Genetic Algorithms}, {Morgan Kaufmann
  Publishers Inc.}, {San Francisco, CA, US}, 1993, p. 416–423.

\bibitem{Back2000}
T.~{B\"ack}, D.~B. Fogel, Z.~Michalewicz, {Evolutionary Computation 2: Advanced
  Algorithms and Operators}, {CRC Press}, {Boca Raton, FL, US}, 2000.

\bibitem{Baker1987}
J.~E. Baker, {Reducing Bias and Inefficiency in the Selection Algorithm}, in:
  {Proceedings of the Second International Conference on Genetic Algorithms and
  Their Application}, {L. Erlbaum Associates Inc.}, {Hillsdale, NJ, US}, 1987,
  p. 14–21.

\bibitem{Kolmogorov1950}
A.~N. Kolmogorov, {Foundations of the Theory of Probability}, {Chelsea
  Publishing Company}, {New York, NY, US}, 1950.

\bibitem{Aarts1989}
E.~H.~L. Aarts, A.~E. Eiben, K.~M. van Hee, {A general theory of genetic
  algorithms}, Tech. rep., {Technische Universiteit Eindhoven}, {Eindhoven,
  Netherlands} (1989).

\bibitem{Alba2008}
E.~Alba, B.~Dorronsoro, {Cellular Genetic Algorithms}, {Operations
  Research/Computer Science Interfaces Series}, {Springer}, {Boston, MA, US},
  2008.
\newblock \href {https://doi.org/10.1007/978-0-387-77610-1}
  {\path{doi:10.1007/978-0-387-77610-1}}.

\bibitem{vonNeumann1966}
J.~von Neumann, A.~W. Burks, {Theory of self-reproducing automata}, {University
  of Illinois Press}, {Champaign, IL, US}, 1966.

\bibitem{Cerruti2020}
U.~Cerruti, S.~Dutto, N.~Murru, {A symbiosis between cellular automata and
  genetic algorithms}, {Chaos, Solitons \&{} Fractals} 134 (2020) 109719.
\newblock \href {https://doi.org/10.1016/j.chaos.2020.109719}
  {\path{doi:10.1016/j.chaos.2020.109719}}.

\bibitem{Peterson1966}
R.~J. Duffin, E.~L. Peterson, {Duality Theory for Geometric Programming}, {SIAM
  Journal on Applied Mathematics} 14~(6) (1966) 1307--1349.
\newblock \href {https://doi.org/10.1137/0114105} {\path{doi:10.1137/0114105}}.

\bibitem{Ackley1987}
D.~H. Ackley, {A connectionist machine for genetic hillclimbing}, {Springer},
  {Boston, MA, US}, 1987.
\newblock \href {https://doi.org/10.1007/978-1-4613-1997-9}
  {\path{doi:10.1007/978-1-4613-1997-9}}.

\bibitem{AluffiPentini1985a}
F.~Aluffi-Pentini, V.~Parisi, F.~Zirilli, {A global optimization algorithm
  using stochastic differential equations}, Tech. Rep.~{\#2791}, {University of
  Wisconsin-Madison, Mathematics Research Center}, {Madison, WI, US} (1985).

\bibitem{AluffiPentini1985b}
F.~Aluffi-Pentini, V.~Parisi, F.~Zirilli, {Sigma -- A stochastic-integration
  global minimization algorithm}, Tech. Rep.~{\#2806}, {University of
  Wisconsin-Madison, Mathematics Research Center}, {Madison, WI, US} (1985).

\bibitem{Easom1990}
E.~E. Easom, {A survey of global optimization techniques}, Master's thesis,
  {University of Louisville}, {Louisville, KY, US} (1990).

\bibitem{Goldstein1971}
A.~A. Goldstein, I.~F. Price, {On descent from local minima}, {Mathematics of
  Computation} 25~(115) (1971) 569--574.
\newblock \href {https://doi.org/10.1090/S0025-5718-1971-0312365-X}
  {\path{doi:10.1090/S0025-5718-1971-0312365-X}}.

\bibitem{Rosenbrock1960}
H.~H. Rosenbrock, {An Automatic Method for Finding the Greatest or Least Value
  of a Function}, {The Computer Journal} 3~(3) (1960) 175--184.
\newblock \href {https://doi.org/10.1093/comjnl/3.3.175}
  {\path{doi:10.1093/comjnl/3.3.175}}.

\bibitem{Schwefel1981}
H.-P. Schwefel, {Numerical optimization of computer models}, {John Wiley \&{}
  Sons, Inc.}, 1981.

\bibitem{Jamil2013}
M.~Jamil, X.-S. Yang, H.-J. Zepernick, {Test Functions for Global Optimization:
  A Comprehensive Survey}, in: {Swarm Intelligence and Bio-Inspired
  Computation}, {Elsevier}, {Oxford, England, Great Britain}, 2013, pp.
  193--222.
\newblock \href {https://doi.org/10.1016/B978-0-12-405163-8.00008-9}
  {\path{doi:10.1016/B978-0-12-405163-8.00008-9}}.

\bibitem{Jamil2013a}
M.~Jamil, X.-S. Yang, {A literature survey of benchmark functions for global
  optimisation problems}, {International Journal of Mathematical Modelling and
  Numerical Optimisation} 4~(2) (2013) 150--194.
\newblock \href {https://doi.org/10.1504/IJMMNO.2013.055204}
  {\path{doi:10.1504/IJMMNO.2013.055204}}.

\bibitem{Grefenstette1986}
J.~J. Grefenstette, {Optimization of Control Parameters for Genetic
  Algorithms}, {IEEE Transactions on Systems, Man, and Cybernetics} 16~(1)
  (1986) 122--128.
\newblock \href {https://doi.org/10.1109/TSMC.1986.289288}
  {\path{doi:10.1109/TSMC.1986.289288}}.

\bibitem{Birattari2010}
M.~Birattari, Z.~Yuan, P.~Balaprakash, T.~{St\"utzle}, {F-Race and Iterated
  F-Race: An Overview}, in: {Experimental Methods for the Analysis of
  Optimization Algorithms}, {Springer}, {Berlin, Heidelberg, Germany}, 2010,
  pp. 311--336.
\newblock \href {https://doi.org/10.1007/978-3-642-02538-9_13}
  {\path{doi:10.1007/978-3-642-02538-9_13}}.

\bibitem{Craenen}
B.~G.~W. Craenen, {Solving Constraint Satisfaction Problems with Evolutionary
  Algorithms}, Ph.D. thesis, {Vrije Universiteit Amsterdam}, {Amsterdam,
  Netherlands} (2005).

\bibitem{Wolpert1997}
D.~Wolpert, W.~Macready, {No free lunch theorems for optimization}, {IEEE
  Transactions on Evolutionary Computation} 1~(1) (1997) 67--82.
\newblock \href {https://doi.org/10.1109/4235.585893}
  {\path{doi:10.1109/4235.585893}}.

\bibitem{Katoch2021}
S.~Katoch, S.~S. Chauhan, V.~Kumar, {A review on genetic algorithm: past,
  present, and future}, {Multimedia Tools and Applications} 80 (2021)
  8091--8126.
\newblock \href {https://doi.org/10.1007/s11042-020-10139-6}
  {\path{doi:10.1007/s11042-020-10139-6}}.

\bibitem{Ghaheri2015}
A.~Ghaheri, S.~Shoar, M.~Naderan, S.~S. Hoseini, {The Applications of Genetic
  Algorithms in Medicine}, {Oman Medical Journal} 30~(6) (2015) 406--416.
\newblock \href {https://doi.org/10.5001/omj.2015.82}
  {\path{doi:10.5001/omj.2015.82}}.

\bibitem{Goudos2016}
S.~K. Goudos, C.~Kalialakis, R.~Mittra, {Evolutionary Algorithms Applied to
  Antennas and Propagation: A Review of State of the Art}, {International
  Journal of Antennas and Propagation} 2016 (2016) 1010459.
\newblock \href {https://doi.org/10.1155/2016/1010459}
  {\path{doi:10.1155/2016/1010459}}.

\bibitem{Kudjo2017}
P.~K. Kudjo, E.~N.~N. Ocquaye, W.~Ametepe, {Review of Genetic Algorithm and
  Application in Software Testing}, {International Journal of Computer
  Applications} 160~(2) (2017) 1--6.
\newblock \href {https://doi.org/10.5120/ijca2017912965}
  {\path{doi:10.5120/ijca2017912965}}.

\bibitem{Lee2018}
C.~K.~H. Lee, {A review of applications of genetic algorithms in operations
  management}, {Engineering Applications of Artificial Intelligence} 76 (2018)
  1--12.
\newblock \href {https://doi.org/10.1016/j.engappai.2018.08.011}
  {\path{doi:10.1016/j.engappai.2018.08.011}}.

\bibitem{Drachal2021}
K.~Drachal, M.~Pawłowski, {A Review of the Applications of Genetic Algorithms
  to Forecasting Prices of Commodities}, {Economies} 9~(1) (2021).
\newblock \href {https://doi.org/10.3390/economies9010006}
  {\path{doi:10.3390/economies9010006}}.

\end{thebibliography}
\end{multicols}

\end{document}